\pdfoutput=1
\documentclass[11pt]{article}

\usepackage{times}

\usepackage{soul}
\usepackage{url}
\usepackage{graphicx}
\usepackage{amsmath}
\usepackage{booktabs}
\usepackage{multirow}
\usepackage{acl}
\usepackage{times}
\usepackage{latexsym}
\usepackage[T1]{fontenc}
\usepackage[utf8]{inputenc}
\usepackage{microtype}

\usepackage{algorithm}
\usepackage{algorithmic}

\usepackage{graphicx}               
\usepackage{tabularx}               
\usepackage{booktabs}
\usepackage{amsmath}
\usepackage{stmaryrd}
\usepackage{xcolor}
\usepackage{soul}
\usepackage{array}

\newcolumntype{C}{>{\centering\arraybackslash}X}
\usepackage{multirow}               
\usepackage{diagbox}                
\usepackage{hhline}                 
\usepackage{amsmath}                
\usepackage{amssymb}                
\usepackage{mathtools}              
\usepackage{enumitem}               
\usepackage[toc,page]{appendix}

\usepackage{subfigure}
\usepackage{booktabs}
\usepackage{extarrows}
\usepackage{makecell}

\definecolor{RoseQuartzBg}{HTML}{F7CAC9}
\definecolor{RoseQuartz}{HTML}{F5A798}
\definecolor{Serenity}{HTML}{92A8D1}
\definecolor{OrangeRed}{rgb}{1.0, 0.27, 0.0}
\definecolor{Red}{rgb}{1.0, 0.0, 0.0}
\definecolor{Turquoise}{HTML}{0F4C81}
\usepackage{xparse}
\NewDocumentCommand{\lifu}{ mO{} }{\textcolor{OrangeRed}{\textsuperscript{\textit{Lifu}}\textsf{\textbf{\small[#1]}}}}
\NewDocumentCommand{\shuaicheng}{ mO{} }{\textcolor{blue}{\textsuperscript{\textit{Shuaicheng}}\textsf{\textbf{\small[#1]}}}}
\NewDocumentCommand{\qn}{ mO{} }{\textcolor{Red}{\textsuperscript{\textit{QN}}\small[#1]}}

\newcommand{\matres}{\textsc{Matres}}
\newcommand{\tbd}{\textsc{TB-Dense}}
\newcommand{\tmprel}[1]{\textit{#1}}
\newcommand{\this}{\textsc{SGT}} 
\usepackage{newfloat}
\usepackage{listings}
\floatstyle{ruled}
\newfloat{listing}{tb}{lst}{}
\floatname{listing}{Listing}
\begin{document}
\title{Extracting Temporal Event Relation with Syntax-guided Graph Transformer} 

\author{Shuaicheng Zhang$^{\clubsuit}$, \  Qiang Ning$^{\spadesuit}$, \ Lifu Huang$^{\clubsuit}$
\\
  $^{\clubsuit}$Virginia Tech, \
  $^{\spadesuit}$Amazon 
 \\
  $^{\clubsuit}${\tt \{zshuai8,lifuh\}@vt.edu}, \
  $^{\spadesuit}${\tt qning@amazon.com}
  }


\maketitle

\begin{abstract}
Extracting temporal relations (e.g., before, after, and simultaneous) among events is crucial to natural language understanding. One of the key challenges of this problem is that when the events of interest are far away in text, the context in-between often becomes complicated, making it challenging to resolve the temporal relationship between them. This paper thus proposes a new Syntax-guided Graph Transformer network (SGT) to mitigate this issue, by (1) explicitly exploiting the connection between two events based on their dependency parsing trees, and (2) automatically locating temporal cues between two events via a novel syntax-guided attention mechanism. Experiments on two benchmark datasets, \matres{} and \tbd{}, show that our approach significantly outperforms previous state-of-the-art methods on both end-to-end temporal relation extraction and temporal relation classification; This improvement also proves to be robust on the contrast set of \matres{}. The code is publicly available at \url{https://github.com/VT-NLP/Syntax-Guided-Graph-Transformer}.

\end{abstract}

\section{Introduction}
\label{lab:intro}

Temporal relationship, e.g., \tmprel{Before}, \tmprel{After}, and \tmprel{Simultaneous}, is important for understanding the process of complex events and reasoning over them. Extracting temporal relationship automatically from text is thus an important component in many downstream applications, such as summarization~\cite{jiang2011natural,ng2014exploiting}, dialog understanding and generation~\cite{ritter2010unsupervised,shi2019unsupervised}, reading comprehension~\cite{harabagiu2005question,sun2018reading,ning2020torque,huang2019cosmos} and future event prediction~\cite{li2021future,lin2022inferring}. While event mentions can often be detected reasonably well~\cite{lin2020joint,huang2020semi,wang2021query,wang2022art}, extracting event-event relationships, especially temporal relationship, still remains challenging~\cite{CZNLJMR21}.



\begin{figure}
\begin{center}
\includegraphics[width=1\linewidth]{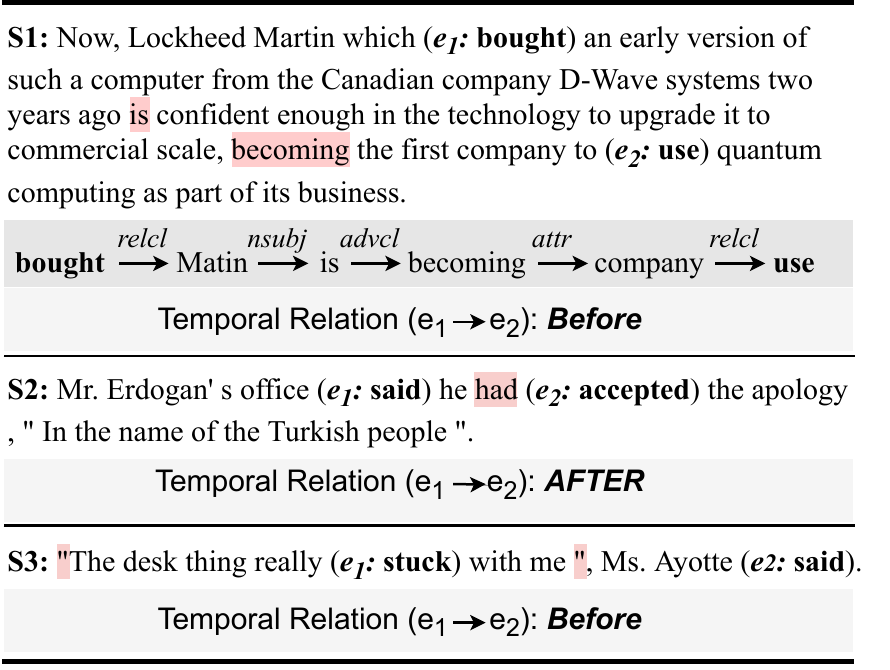}
\caption{Examples of temporal relation annotations. Event mentions are boldfaced, the temporal relations between these events are listed below each sentence, and the temporal cues deciding those temporal relations are highlighted in red.
}
\label{Fig:Temporal_Relation_Cues}
\end{center}
\end{figure}

%

 \begin{figure*}[!ht]
\centering
  \includegraphics[width=1\textwidth,keepaspectratio]{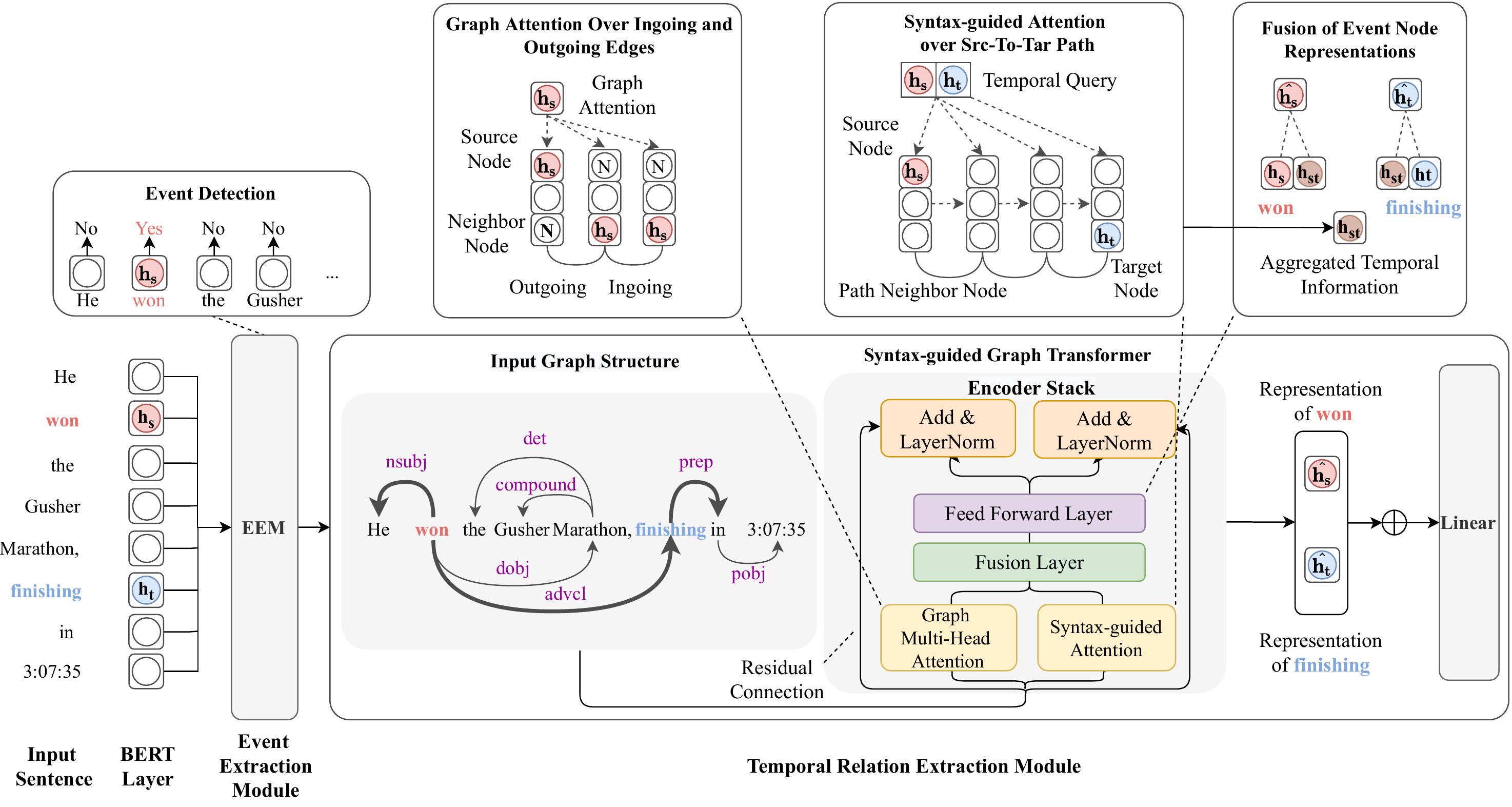}
  \caption{Architecture overview. The tokens highlighted with red and blue colors in the Input Sentence show the source and target events to be detected. The bold edges in the Input Graph Structure indicate the triples from the dependency path between the source and target event mentions as well as their surrounding context, and are considered by the syntax-guided attention.
  }
\label{fig:flow}
\end{figure*}

Recent studies~\cite{han2019joint,ning2019structured,vashishtha2019fine,wang2020joint} have shown improved performance in temporal relation extraction 
by leveraging the contextual representations learned from pre-trained language models~\cite{devlin2018bert,liu2019roberta}. 
However, one remaining challenge of this task is that it requires accurate characterization of the connection between two event mentions and the cues indicating their temporal relationship, especially when the context is wide and complicated. For instance,
by manually examining 200 examples of human annotated temporal relations from the \matres~\cite{NingWuRo18} dataset, we find that about 52\% of the temporal cues\footnote{Temporal cues refer to the words of which the semantic meaning or related syntactic relations can determine the temporal relation of two event mentions. } come from the connection between two event mentions (e.g., S1 in Fig.~\ref{Fig:Temporal_Relation_Cues}), 39\% from their surrounding contexts (S2 in Fig.~\ref{Fig:Temporal_Relation_Cues}) and the remaining 9\% from others, e.g., event co-reference or subordinate clause structures (S3 in Fig.~\ref{Fig:Temporal_Relation_Cues}). 

Syntactic features, such as dependency parsing trees, have proved to be effective in characterizing the connection of two event mentions in pre-neural methods~\cite{chambers-2013-navytime,chambers2014dense,mirza2016catena}. 
However, how to make use of these features has been under-explored since the adoption of neural methods in this field.
This paper closes this gap with a novel Syntax-guided Graph Transformer (\this{}) network -- in addition to the attention heads in a typical Graph Transformer, we bring in a new attention mechanism that specifically looks at the path from a source node to a target node over dependency parsing trees. \this{} thus not only learns event representations as in a typical Graph Transformer, but also provides a way to represent syntactic dependency information between a pair of events (for temporal relation extraction, this means attending to the aforementioned temporal cues). We conduct experiments on two benchmark datasets, \matres~\cite{NingWuRo18} and \tbd~\cite{cassidy2014annotation} on both end-to-end temporal relation extraction and classification, which demonstrate the effectiveness of \this{} over previous state-of-the-art methods. 
Experiments on the contrast set~\cite{Gardner2020Evaluating} of \matres{} further proves the robustness of our approach.
\section{Approach}

Figure~\ref{fig:flow} shows the overview of our approach. Given an input sentence $\tilde{s} = [w_1, w_2, ..., w_n]$ with $n$ tokens, we aim to detect a set of event mentions $\{e_1, e_2, ...\}$ where each event mention $e_i$ may contain one or multiple tokens by leveraging the contextual representations learned from a pre-trained BERT~\cite{devlin2018bert} encoder. Then, following previous studies~\cite{ning2019improved,ning2019structured,han2019joint,wang2020joint}, we consider each pair of event mentions that are detected from one or two continuous sentences, and predict their temporal relationship. 

To effectively capture the temporal cues between two event mentions, we build a dependency graph from one or two input sentences 
and design a new Syntax-guided Graph Transformer network to automatically learn a new contextual representation for each event mention by considering the triples that they are locally involved as well as the triples along the dependency path of the two event mentions within the dependency graph. Finally, the two event mention representations are concatenated to predict their temporal relationship.


 
\subsection{Sequence Encoder}
Given an input sentence $\tilde{s} = [w_1, w_2, ..., w_n]$, we apply the same tokenizer as BERT~\cite{devlin2018bert} to get all the subtokens. Then, we feed the sequence of subtokens as input to a pre-trained BERT model to get a contextual representation for each token $w_i$. If a token $w_i$ is split into multiple subtokens, we use the contextual representation of the first subtoken to represent $w_i$. To enrich the contextualized representations, for each token, we create a one-hot Part-of-Speech (POS) tag vector and concatenate it with BERT contextual embeddings. In this way, we obtain a final representation $\boldsymbol{c}_i$\footnote{We use bold lower case symbols to denote vectors.} for each $w_i$. These representations will be later used for event mention detection and also as the initial representations to our syntax-guided graph transformer network.

\subsection{Event Detection}
To detect event mentions from the sentence, we take the contextual representation of each word as input to a binary linear classifier to determine whether it is an event mention or not, which is optimized by minimizing the following binary cross-entropy loss: 
\begin{align*}
    \small
    &\tilde{\boldsymbol{y}}_{i} =  \text{softmax}(\boldsymbol{W}_{eve}\boldsymbol{c}_i + \boldsymbol{b}_{eve}) 
    \\
    &\mathcal{L}_{eve} = - \sum_{\tilde{s}\in\mathcal{S}}\sum_{i = 1}^{|\tilde{s}|}\sum_{\pi\in\{0, 1\}} \alpha_{\pi} y_{i,\pi} \log (\tilde{y}_{i,\pi})
\end{align*}
where $\mathcal{L}_{eve}$ denotes the cross-entropy loss for event detection. 
$\mathcal{S}$ is the set of sentences in the training dataset. 
$\alpha_{\pi}$ is a weight coefficient for each class (0 or 1) to mitigate the data imbalance problem and $\alpha_{0} + \alpha_{1}=1$. $y_{i,\pi}$ is a binary indicator to show whether $\pi$ is the same as the groundtruth binary label ($y_{i,\pi}=1$) or not ($y_{i,\pi}=0$). $\tilde{y}_{i, \pi}$ denotes the probability of the $i$-th token in $s$ being predicted with a binary class label $\pi$. $\boldsymbol{W}_{eve}$ and $\boldsymbol{b}_{eve}$ are learnable parameters.

\subsection{Syntax-guided Graph Transformer}
From the example sentences in Fig.~\ref{Fig:Temporal_Relation_Cues}, the temporal cues for characterizing the temporal relationship between two event mentions mainly come from their surrounding contexts as well as their connections from their syntactic dependency path. 
However, a sequence encoder usually fails to capture such information, especially when the context between two event mentions is complicated, thus we further design a new Syntax-guided Graph Transformer (\this{}) network.

Given a source event $e_s$ and a target event $e_t$ detected from one or two continuous sentences, we apply a public dependency parser\footnote{\url{https://spacy.io/api/dependencyparser}} to parse each sentence into a tree-graph and connect the graphs of two continuous sentences with an arbitrary \textit{cross-sentence} edge~\cite{peng2017cross,cheng2017classifying} pointing from the root node of the preceding sentence to the root node of the following one, and obtain a graph $G = (V, E).$
For each node $v_i$, we use $\mathcal{N}_i^{in}=\{(v_k,r_{ki},v_i)\in E | v_k, v_i \in V\}$ and $\mathcal{N}^{out}_{i}=\{(v_i,r_{ij},v_j)\in E | v_i, v_j \in V\}$ to denote all the neighbor triples of $v_i$ with in-going and out-going edges respectively, $r\in\Upsilon$ where $\Upsilon$ is the label set for syntactic dependency relation, and use $\mathcal{P}_{ij} = \{(v_i, r_{ig}, v_g), ..., (v_h, r_{hj},v_j)\}$ as the triple set along the path from $v_i$ to $v_j$.

\paragraph{Node Representation Initialization}

For each node $v_i$ in graph $G$, we map it to a particular token $w_{i^{'}}$ from the original sentence and obtain a contextual representation $\boldsymbol{c}_{i^{'}}$ from the BERT encoder. Then, we learn an initial node representation for each node $v_i$ as: 
\begin{align*}
    \boldsymbol{h}^{0}_{i} = \boldsymbol{W}_e\boldsymbol{c}_{i^{'}} + \boldsymbol{b}_e
\end{align*}
where $\boldsymbol{W}_e$ and $\boldsymbol{b}_e$ are learnable parameters.

\paragraph{Graph Multi-head Self-attention} 
Following transformer model~\cite{vaswani2017attention,wang2020amr}, we adapt the multi-head self-attention to learn a contextual representation for each node in the graph $G$. 
Each node $v_i$ in graph $G$ is associated with a set of neighbor triples 
$\mathcal{N}^{in}_{i}\cup\mathcal{N}^{out}_{i}$
and a node representation $\boldsymbol{h}_i^{l-1}$ where $l$ is the index of a layer in our transformer architecture. 
To perform self-attention, we first apply a linear transformation to obtain a query vector based on each node $v_i$, and employ another two linear transformations to get the key and value vectors based on the node's neighbor triples:
\begin{align*}
\small
   & \boldsymbol{Q}^l_{i} = \boldsymbol{W}_q^m \boldsymbol{h}^{l-1}_i
   \\
   &\boldsymbol{K}^l_{ij} = \boldsymbol{W}_k^m \boldsymbol{R}^{l-1}_{ij}
    \\
    &\boldsymbol{U}^l_{ij} = \boldsymbol{W}_u^m \boldsymbol{R}^{l-1}_{ij}
    \\&\boldsymbol{R}^{l-1}_{ij} = \boldsymbol{W}_r^m ( \boldsymbol{h}^{l-1}_{i}\bigparallel \boldsymbol{r}_{ij} \bigparallel \boldsymbol{h}^{l-1}_{j}) + \boldsymbol{b}_r^m
\end{align*}
where $m$ is the index of a particular head. $\boldsymbol{Q}^l_{i}$ denotes a query vector corresponding to node $v_i$, $\boldsymbol{K}^l_{ij}$ and $\boldsymbol{U}^l_{ij}$ is a key and value vector respectively, and both of them are learned from a triple $(v_i, r_{ij}, v_j)\in \mathcal{N}^{in}_{i}\cup\mathcal{N}^{out}_{i}$, which is represented as $\boldsymbol{R}_{ij}$. $m$ is the index of a particular head. $\bigparallel$ denotes the concatenation operation. $\boldsymbol{r}_{ij}$ denotes the representation of a particular relation $r_{ij}$ between $v_i$ and $v_j$, which is randomly initialized and optimized by the model. 
$\boldsymbol{W}_{q}^m$, $\boldsymbol{W}_{k}^m$, $\boldsymbol{W}_{u}^m$, $\boldsymbol{W}_r^m$ and $\boldsymbol{b}_r^m$ are learnable parameters. 

For each node $v_i$, we then perform self-attention over all the neighbor triples that it is involved, and compute a new context representation with multiple attention heads:
\begin{align*}
\small
    &\boldsymbol{g}_i^l =  (\bigparallel^{M}_{m}\boldsymbol{\text{Head}}^{m}_{i})\boldsymbol{W}_o
    \\
    &\boldsymbol{\text{Head}}^m_{i} = \text{softmax}(\frac{\boldsymbol{Q}^{l}_i(\boldsymbol{K}^{l})^{\top}}{\sqrt{d_k}})\boldsymbol{U}^{l}
\end{align*}
where $\boldsymbol{g}_i^l$ is the aggregated representation computed over all neighbor triples of node $v_i$ with $M$ attention heads at $l$-th layer. $\boldsymbol{g}_i^l$ will be later used to learn the updated representation of node $v_i$.  
$\sqrt{d_k}$ is the scaling factor denoting the dimension size of each key vector. $\boldsymbol{W}_o$ is a learnable parameter.

\paragraph{Syntax-guided Attention} To automatically find the indicative temporal cues for two event mentions from their connection as well as surrounding contexts, we design a new syntax-guided attention mechanism.  
For two event nodes $v_s$ and $v_t$, we first extract the set of nodes from the dependency path between $v_s$ and $v_t$ (including $v_s$ and $v_t$), which is denoted as $\Theta_{st}$. We then get all the triples from the dependency path between $v_s$ and $v_t$ as well as the triples that any node from $\Theta_{st}$ is involved, which are denoted as $\Phi_{st} = \cup_{v_i \in \Theta_{st}}\{\mathcal{N}^{in}_{i} \cup \mathcal{N}^{out}_{i}\}\cup\mathcal{P}_{st}$. To compute the syntax-guided attention over all the triples from $\Phi_{st}$, we apply three linear transformations to get the query, key and value vectors where the query vector is obtained from the representation of two event mentions, and key and value vectors are computed from the triples in $\Phi_{st}$:
\begin{align*}
\small
   & \tilde{\boldsymbol{Q}}^l_{st} = \tilde{\boldsymbol{W}}_{q}^m\cdot(\boldsymbol{h}^{l-1}_s\bigparallel \boldsymbol{h}^{l-1}_t)^x
   \\
    &\tilde{\boldsymbol{K}}^l_{ij} = \tilde{\boldsymbol{W}}_k^m \tilde{\boldsymbol{R}}^{l-1}_{ij}
    \\
    &\tilde{\boldsymbol{U}}^l_{ij} = \tilde{\boldsymbol{W}}_{u}^m \tilde{\boldsymbol{R}}^{l-1}_{ij}
    \\
    & \tilde{\boldsymbol{R}}^{l-1}_{ij} = \tilde{\boldsymbol{W}}_r^m ( \boldsymbol{h}^{l-1}_{i}\bigparallel \boldsymbol{r}_{ij} \bigparallel  \boldsymbol{h}^{l-1}_{j}) + \tilde{\boldsymbol{b}}_r
\end{align*}
where $m$ is the index of a particular head, $\tilde{\boldsymbol{Q}}^l_{st}, \tilde{\boldsymbol{K}}^l_{ij}, \tilde{\boldsymbol{U}}^l_{ij}$ denote the query, key and value vectors respectively. $\tilde{\boldsymbol{R}}^{l-1}_{ij}$ is the representation of a triple $(v_i, r_{ij}, v_j)\in\Phi_{st}$. $\tilde{\boldsymbol{W}}_{q}^m$, $\tilde{\boldsymbol{W}}_{k}^m$, $\tilde{\boldsymbol{W}}_{v}^m$ and $\tilde{\boldsymbol{W}}_r^m$ are learnable parameters. 


Given the query vector, we then compute the attention distribution over all triples from $\Phi_{st}$ and get an aggregated representation to denote the meaningful temporal features captured from the connection between two event mentions and their surrounding contexts.
\begin{align*}
\small
    &\tilde{\boldsymbol{g}}_{st}^l =  (\bigparallel^{M}_{m} \tilde{\text{Head}}^{m}_{st}) \cdot\tilde{\boldsymbol{W}}_p \\
    &
    \tilde{\text{Head}}^{m}_{st} =\text{softmax}(\frac{\tilde{\boldsymbol{Q}}^{l}_{st}(\tilde{\boldsymbol{K}}^{l})^{\top}}{\sqrt{d_k}})\cdot\tilde{\boldsymbol{U}}^l
\end{align*}
where $\tilde{\boldsymbol{g}}_{st}^l$ is the aggregated temporal related information from all the triples in $\Phi_{st}$ based on the syntax-guided attention at $l$-th layer. $\boldsymbol{W}_p$ is a learnable parameter.

\paragraph{Node Representation Fusion} Each event node in graph $G$ will receive two representations learned from the multi-head self-attention and syntax-guided attention, thus we further fuse the two representations for both the source node $v_s$ and the target node $v_t$: 
\begin{align*}
\small
    &\boldsymbol{\hat{h}}_{s}^l = \tilde{\boldsymbol{W}}_{f} (\boldsymbol{g}_{s}^{l} \bigparallel \tilde{\boldsymbol{g}}_{st}^l)\;, \;\;
    \boldsymbol{\hat{h}}_{t}^l = \tilde{\boldsymbol{W}}_{f} (\tilde{\boldsymbol{g}}_{st}^l \bigparallel \boldsymbol{g}_{t}^{l})
\end{align*}
where $\boldsymbol{g}_{s}^l$ and $\boldsymbol{g}_{t}^l$ denote the context representations learned from the multi-head self-attention for $v_s$ and $v_t$. $\tilde{\boldsymbol{g}}_{st}^l$ denotes the representation learned from the triples from $\Phi_{st}$ using syntax-guided attention. $\boldsymbol{\hat{h}}_{s}^l$ and $\boldsymbol{\hat{h}}_{t}^l$ are the fused representations of $v_s$ and $v_t$, respectively. $\tilde{\boldsymbol{W}}_{f}$ is a learnable parameter.

For each non-event node $v_i$, which only receives a context representation $\boldsymbol{g}_i^l$ learned from the multi-head self-attention, we apply a linear projection and get a new node representation:
\begin{align*}
\small
    \boldsymbol{\hat{h}}_{i}^l = \boldsymbol{W}_t \boldsymbol{g}_i^l
\end{align*}

Our Syntax-guided Graph Transformer encoder is composed of a stack of multiple layers, while each layer consists of the two attention mechanisms and the fusion sub-layer. We use residual connection followed by LayerNorm for each layer to get the final representations of all the nodes:
\begin{align*}
\small
    \boldsymbol{H}^l = \text{LayerNorm}(\boldsymbol{\hat{H}}^l + \boldsymbol{H}^{l-1})
\end{align*}


\subsection{Temporal Relation Prediction}

To predict the temporal relation between two event mentions $e_s$ and $e_t$, we concatenate the final hidden states of $v_s$ and $v_t$ obtained from the Syntax-guided Graph Transformer network, and apply a Feedforward Neural Network (FNN) to predict their relationship
\begin{align*}
\tilde{\boldsymbol{y}}_{st} = \text{softmax}(\boldsymbol{W}_{z}(\boldsymbol{h}^L_{s} \bigparallel \boldsymbol{h}^L_{t}) + \boldsymbol{b}_t)
\end{align*}
where 
$\tilde{\boldsymbol y}_{st}$ denotes the probabilities over all possible temporal relations between event mentions $e_s$ and $e_t$. 

The training objective is to minimize the following cross-entropy loss function:
\begin{align*}
\small
    \mathcal{L}_{rel} = -\sum_{st\in\Delta}\sum_{x\in X} \beta_xy_{st,x} \text{log}(\tilde{y}_{st,x}))
\end{align*}
where $\Delta$ denotes the total set of event pairs for temporal relation prediction and $X$ denotes the whole set of relation labels. $y_{st,x}$ is a binary indicator (0 or 1) to show whether $x$ is the same as the groundtruth label ($y_{st,x}=1$) or not ($y_{st,x}=0$). We also assign a weight $\beta_x$ to each class to mitigate the label imbalance issue. 

\section{Experiment}



\subsection{Experimental Setup}
We perform experiments on two public benchmark datasets for temporal relation extraction: (1) \tbd~\cite{cassidy2014annotation}, which is a densely annotated dataset with 6 types of relations: \textit{Before}, \textit{After}, \textit{Simultaneous}, \textit{Includes}, \textit{Is\_included} and \textit{Vague}. (2) \matres~\cite{NingWuRo18}, 
which annotates verb event mentions along with 4 types of temporal relations: \textit{Before}, \textit{After}, \textit{Simultaneous} and \textit{Vague}. Additionally, we use POS tag information from~\matres{} provided by~\cite{ning2019improved}. For~\tbd, we use spacy annotation for predicting POS tag information which is based on Universal POS tag set\footnote{\url{https://spacy.io/api/data-formats}}.  
For both benchmark datasets, we use the same train/dev/test splits as previous studies~\cite{ning2019improved,ning2019structured,han2019deep,han2019joint}. Note that, for evaluation, similar as previous work, we disregard the \textit{Vague} relation from both datasets (in the evaluation phase, we simply remove all ground truth \textit{Vague} relation pairs).  In addition, we will only consider event pairs from adjacent sentences due to the fact that it will require an exponential number of annotations if we also consider event pairs from non-adjacent sentences, which is beyond the scope of this study. Table~\ref{tab:corpora stats} shows statistics of the two datasets and Table~\ref{tab:label_distribution} shows the label distribution. 

\begin{table}[h]
\small
\centering
\begin{tabular}{ccccc}
\toprule
\multicolumn{2}{c}{Corpora} & Train & Dev & Test \\ 
\midrule
\multicolumn{1}{c|}{\multirow{2}{*}{\tbd}} & \multicolumn{1}{c|}{\# Documents} & \multicolumn{1}{c|}{22} & \multicolumn{1}{c|}{5} & 9 \\
\multicolumn{1}{c|}{} & \multicolumn{1}{c|}{\# Relation Pairs} & \multicolumn{1}{c|}{4,032} & \multicolumn{1}{c|}{629} & 1,427 \\ \midrule 
\multicolumn{1}{c|}{\multirow{2}{*}{\matres}} & \multicolumn{1}{c|}{\# Documents} & \multicolumn{1}{c|}{255} & \multicolumn{1}{c|}{20} & 25 \\
\multicolumn{1}{c|}{} & \multicolumn{1}{c|}{\# Relation Pairs} & \multicolumn{1}{c|}{13K} & \multicolumn{1}{c|}{2.6K} & 837 \\ 
\toprule
\end{tabular}
\caption{Data statistics for \tbd~and \matres}
\label{tab:corpora stats}
\end{table}

\begin{table}[h]
\small
\centering
\begin{tabular}{l|r|r|r|r}
\toprule
Labels & \multicolumn{2}{c|}{\tbd} & \multicolumn{2}{c}{\matres} \\ \midrule
Before & 384 & 26.9\% & 417 & 49.8\% \\
After & 274 & 19.2\% & 266 & 31.8\% \\
Includes & 56 & 3.9\% & - & - \\
Is\_Included & 53 & 3.7\% & - & - \\
Simultaneous & 22 & 1.5\% & 31 & 3.7\% \\
Vague & 638 & 44.7\% & 133 & 15.9\% \\ 
\toprule
\end{tabular}
\caption{Label distribution for \tbd~and \matres. For each dataset, the first column shows the number of instances of each relation type while the second column shows the percentage.}
\label{tab:label_distribution}
\end{table}

\paragraph{Implementation Details} For fair comparisons with previous baseline approaches, we use the pre-trained bert-large-cased model\footnote{\url{https://huggingface.co/transformers/pretrained_models.html}} for fine-tuning and optimize our model with BertAdam. We optimize the parameters with grid search: training epoch 10, learning rate $\in \{3e\text{-}6,1e\text{-}5\}$, training batch size $\in\{16, 32\}$, encoder layer size $\in\{4, 12\}$, number of heads $\in\{1, 8\}$. During training, we first optimize the event extraction module for 5 epochs to warm up, and then jointly optimize both event extraction and temporal relation extraction modules using gold event pairs for another 5 epochs.

\subsection{Results}


\begin{table*}[!tp]
\small
\centering
\scalebox{1}{
\begin{tabular}[width=\linewidth]{c|p{4.7cm}<{\centering}|c|c|c}
\toprule
Dataset & Model & Pre-trained Model & Event Detection & Relation Extraction \\ 
\midrule
\multirow{2}{*}{\tbd} & HNP19~\cite{han2019joint} & BERT Base &  90.9 & 49.4 \\  
 & Our Approach & BERT Base &  \textbf{91.0} & \textbf{51.8} \\ 
\midrule
\multirow{3}{*}{\matres} & CogCompTime2.0~\cite{ning2019improved} & BERT Base & 85.2 & 52.8 \\  
 & HNP19~\cite{han2019joint} & BERT Base &  87.8 & 59.6 \\ \  
 & Our Approach & BERT Base &  \textbf{90.5} & \textbf{62.3} \\ 
 \toprule
\end{tabular}
}
\caption{Comparison of various approaches on joint event and relation extraction with F-score (\%). Note that HPN19 fixes BERT embeddings but relies on BiLSTM to capture the contextual features.}


\label{tab:joint score}  
\end{table*}

\begin{table*}[t]
\small
\centering
\scalebox{1}{
\begin{tabular}{c|c|c|c}
\toprule
\multicolumn{1}{c|}{Dataset} & Model & Pre-trained Model & Relation Classification (F-score \%) \\ \midrule
\multirow{6}{*}{\tbd} 
& LSTM~\cite{cheng2017classifying} & BERT Base & 62.2 \\
& HNP19~\cite{han2019joint} & BERT Base & 64.5 \\
& Our Approach & BERT Base & \textbf{66.7} \\
\cmidrule{2-4}
 & PSL~\cite{zhou2020clinical} & RoBERTa Large & 65.2 \\
 & DEER~\cite{han2020deer} & RoBERTa Large & 66.8 \\
 & Our Approach & BERT Large & \textbf{67.1} \\ \midrule
\multirow{7}{*}{\matres} & CogCompTime2.0~\cite{ning2019improved} & BERT Base & 71.4 \\
& LSTM~\cite{cheng2017classifying} & BERT Base & 73.4 \\
 & HNP19~\cite{han2019joint} & BERT Base & 75.5 \\
 & Our Approach & BERT Base & \textbf{79.3} \\
 \cmidrule{2-4}
 & HMHD20~\cite{wang2020joint} & RoBERTa Large & 78.8 \\
 & DEER~\cite{han2020deer} & RoBERTa Large & 79.3 \\
 & Our Approach & BERT Large & \textbf{80.3} \\ \toprule
\end{tabular}
}
\caption{Comparison of various approaches on temporal relation classification with gold event mentions as input.}
\label{tab:gold event mention}
\end{table*}

We 
evaluate \this{} against two public benchmark datasets under two settings: (1) joint event and temporal relation extraction (Table~\ref{tab:joint score}); (2) temporal relation classification, where the gold event mentions are known beforehand (Table~\ref{tab:gold event mention}). Note in the ``joint'' setting, we adopt the same strategy proposed in \cite{han2019joint}: we first train the event extraction module, and then jointly optimize both event extraction and temporal relation extraction (using gold event pairs as input to ensure training quality) modules.
Overall, we observe that our approach significantly outperforms baseline systems in both settings, with up to 7.9\% absolute F-score gain on~\matres{} and 2.4\% on~\tbd{}. 

From Table~\ref{tab:joint score}, we see that our approach achieves better performance on event detection than baseline methods though they are based on the same BERT encoder. This is possibly because, during joint training, our approach leverages the dependency parsing trees, which improves the contextual representations of the BERT encoder.
In Table~\ref{tab:gold event mention}, unlike other models which are based on larger contextualized embeddings such as RoBERTa, our approach with BERT base achieves comparable performance, and further surpasses the state-of-the-art baseline methods using BERT-large embeddings, which demonstrate the effectiveness of our Syntax-guided Graph Transformer network.

Some studies~\cite{ning2019improved,han2019joint,wang2020joint,zhou2020clinical} focus on resolving the inconsistency in terms of the symmetry and transitivity of the temporal relations. For example, if event A and event B are predicted as \textit{Before}, event B and event C are predicted as \textit{Before}, then if event A and event C are predicted as \textit{Vague} or \textit{After}, it will be considered as inconsistent. However, our approach shows consistent predictions with few inconsistent cases when \textit{Simultaneous} relation is involved. This analysis also demonstrates that our approach can correctly capture the temporal cues between two event mentions.

\begin{table}[tp]
\small
\centering
\scalebox{0.8}{
\begin{tabular}{p{2.8cm}|p{1.6cm}<{\centering}|p{1.6cm}<{\centering}|p{1.6cm}<{\centering}}
\toprule
Model &  Original Test & Contrast & Consistency \\
\midrule
CogCompTime2.0 \cite{ning2019improved} & 73.2 & 63.3  & 40.6 \\
\midrule
Our Approach & \textbf{77.0} & \textbf{64.8} & \textbf{49.8} \\
\toprule
\end{tabular}}
\caption{Evaluation on the contrast set of~\matres. Original Test indicates the accuracy on 100 examples sampled from the original~\matres~test set following~\cite{Gardner2020Evaluating}. Contrast shows the accuracy score on 401 examples perturbed from the original 100 examples. Consistency is defined as the percentage of the original 100 examples for which the model's predictions of the perturbed examples are all correct in the contrast set.}  
\label{tab:contrast}
\end{table}

\begin{figure*}[h]
    \centering
    \includegraphics[width=0.9\textwidth,keepaspectratio]{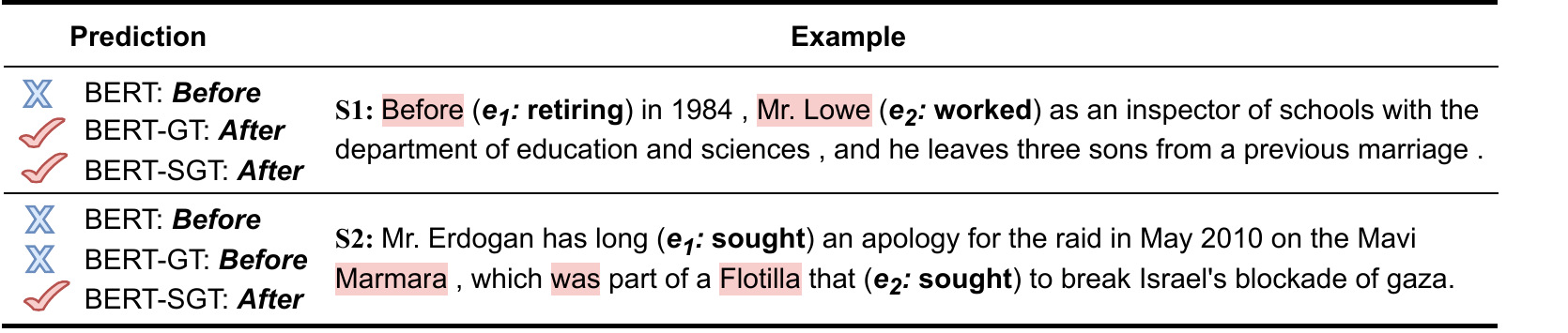}
    \caption{Comparison of the predictions from BERT, BERT-GT and our approach.}
    \label{fig:comparison}
\end{figure*}

We also examine the correctness and robustness of our approach on a contrast set of \matres~\cite{Gardner2020Evaluating}, which is created with small manual perturbation based on the original test set of \matres~in a meaningful way, such as rephrasing the sentence or simply changing a word of the sentence to alter the relation type. The contrast set provides a local view of a model’s decision boundary, thus it can be used to more accurately evaluate a model’s true linguistic capabilities.
Table~\ref{tab:contrast} shows that our approach significantly outperforms the baseline model on both the original test set and the corresponding contrast set. 
The contrast consistency in Table~\ref{tab:contrast} also indicates how well a model's decision boundary aligns with the actual decision boundary of the test instances, based on which we can see that by explicitly capturing temporal cues, our approach is more accurate and robust than the baseline method.

\paragraph{Ablation Study}
We further conduct ablation studies to compare the performance of our approach with two ablated versions of our method: (1) BERT with Graph Transformer (BERT-GT), for which we remove the syntaxic-guided attention and only rely on the standard multi-head self-attention to obtain graph-based contextual representations of two event mentions and then predict their relation; (2) BERT, where we further remove the Graph Transformer, and only use the pre-trained BERT language model to encode the sentence and predict the temporal relationship of two event mentions based on their contextual representations.

\begin{table}[h]
\small
\centering
\begin{tabular}{l|c|c}
\toprule
Ablation & F-score (\%) & Gain (\%) \\ \midrule
BERT-SGT & 79.3 & 0 \\ 
BERT-GT & 77.5 & -2.0 \\ 
BERT & 75.5 & -3.8\\ 
\toprule
\end{tabular}
\caption{Ablation study on \matres. We use BERT base as the comparison basis.}
\label{tab:ablation study}
\end{table}

Table~\ref{tab:ablation study} also shows that by adding Graph Transformer, BERT-GT achieves 2.0\% absolute F-score improvement over the BERT baseline model, demonstrating the benefit of dependency parsing trees to temporal relation prediction. By further adding the new syntax-guided attention into Graph Transformer, the absolute improvement on F-score (1.8\%) shows the effectiveness of our new Syntax-guided Graph Transformer and the importance of capturing temporal cues from the connection of two event mentions as well as their surround contexts.






Figure~\ref{fig:comparison} shows two examples as qualitative analysis. In S1, BERT mistakenly predicts the temporal relation as \textit{Before} probably because it's confused by the context word \textit{Before}. However, by incorporating the dependency graph, especially the triples \{\textit{worked}, \textit{prep}, \textit{Before}\}, \{\textit{Before}, \textit{pcomp}, \textit{retiring}\} and the path between the two event mentions, \textit{worked}$\rightarrow$\textit{prep}$\rightarrow$\textit{Before}$\rightarrow$\textit{pcomp}$\rightarrow$\textit{retiring}, both BERT-GT and our approach correctly determine the relation as \textit{After}. In S2, both BERT and BERT-GT mistakenly predict the temporal relation as \textit{Before} as the context between the two event mentions is very wide and complicated, and these two event mentions are not close within the dependency graph. However, by explicitly considering and understanding the connection between the two event mentions, $\textit{sought}_{e_1}$$\rightarrow$\textit{on}$\rightarrow$\textit{Marmara}$\rightarrow$\textit{was}$\rightarrow$\textit{part}$\rightarrow$\textit{Flotilla}
$\rightarrow$\textit{sought}$_{e_2}$, our approach correctly predicts the temporal relation between the two event mentions.




\subsection{SGT on Temporal Cues}

To analyze the source of temporal cues for relation prediction, we randomly sample 100 correct event relation predictions given gold event mentions from~\matres~and select the triple that has the highest temporal attention weight from the last layer of the Syntax-guided Graph Transformer network as a temporal cue candidate. We manually evaluate the validity of each temporal cue candidate, and further analyze if the cue is from the dependency path between two event mentions, their surrounding context, or both. Our analysis shows that about 64\% of the temporal cues are valid, 37\% of them come from the dependency path, 17\% are from local context, and the remaining 10\% are from both. This verifies our initial observation that most of the temporal cues are from the dependency path between two event mentions as well as their surrounding context. It also demonstrates the effectiveness of our new syntax-guided attention mechanism. 

\subsection{Impact of Wide Context}

\begin{figure}[h!]
    \centering
    \includegraphics[width=1\linewidth, height=8cm, keepaspectratio]{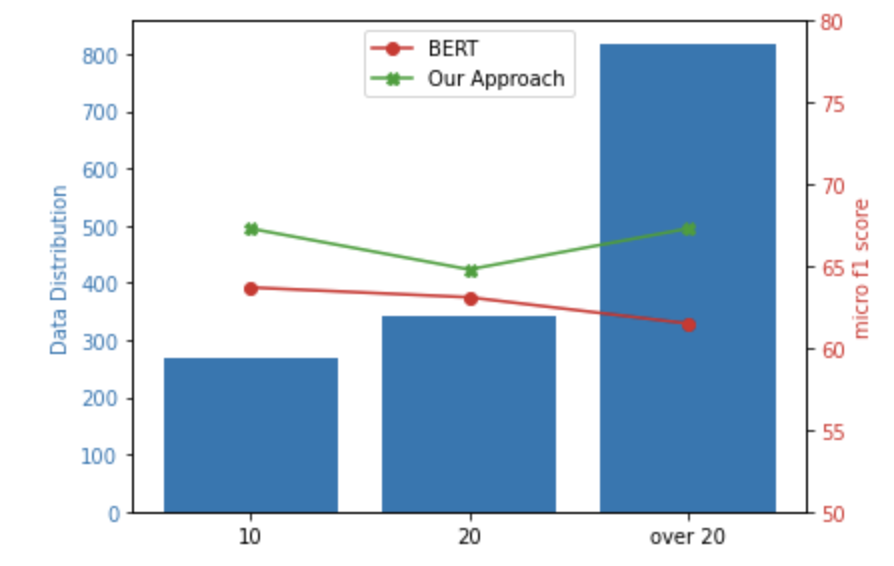}
    \caption{Context width analysis on \tbd. The X axis shows the number of tokens between two events mentions. The left Y axis shows the data distribution of each width category indicating with blue bars. The right Y axis denotes the micro F-score for each width category.}
    \label{fig:context width}
\end{figure}

We further illustrate the impact of context width to both baseline model and our approach. For fair comparison, we use three context width category, [context length $<$ 10, 10 $<$ context length $<$ 20, context length $>$ 20 ]. As we can see in Fig.~\ref{fig:context width}, the first category has 267 pairs, the second category has 343 pairs and the third category has 817 pairs. From our results, we observe that the BERT baseline cannot predict the temporal relation of two event mentions with wide context but rather working well when the event mentions are close to each other. Our model overall performs slightly worse in the second category but in general is very good at predicting the temporal relationship for the event mentions with short and context width. This also proves the benefit of syntactic parsing trees to the prediction of temporal relationship. For the second category where the context length is within [10, 20], the performance of our approach slightly drops due to two reasons: (1) the training samples within this range are not as sufficient as the other two categories; (2) for most event pairs from this category, their dependency path is very
long and there is no explicit temporal indicative features within their context or dependency path, making it more difficult for the model to predict their temporal relationship.

\begin{figure*}[]
    \centering
    \includegraphics[width=\textwidth]{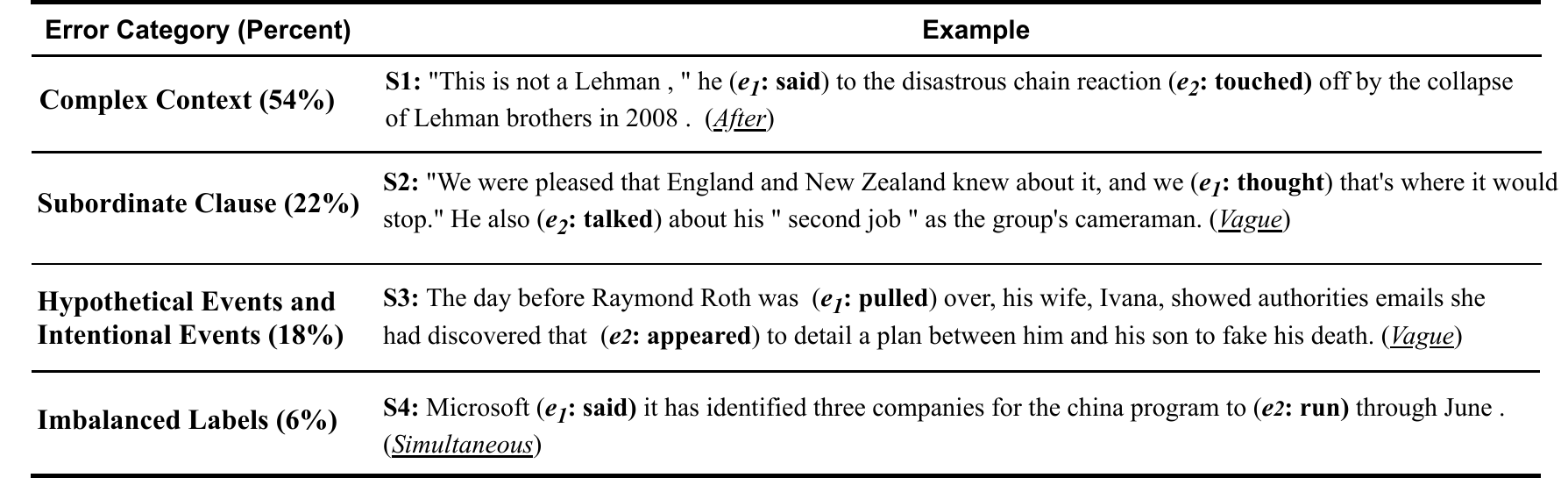}
    \caption{Types of remaining errors}
    \label{fig:errors}
\end{figure*}


\subsection{Remaining Errors}

We randomly sample 100 classification errors from the output of our approach and categorize them into four categories. As Figure~\ref{fig:errors} shows, the first category is due to the complex or ambiguous context (54\% of the total errors). The second category is due to the complicated subordinate clause structure, especially the clauses that are related to quote or reported speech, e.g., S2 in Figure~\ref{fig:errors}. The third error category is that our approach cannot correctly differentiate the actual events from the hypothetical and intentional events, while in most cases, the temporal relation among hypothetical and intentional events is annotated as \textit{Vague}. The last category is due to the lack of sufficient annotation. We observe that none of the \textit{Simultaneous} relation can be correctly predicted for \matres~dataset as the percentage of \textit{Simultaneous} (3.7\%) is much lower than other relation types. In \tbd~dataset, labels are even more imbalanced as the percentage of \textit{Vague} relation is over 50\% while the percentage of \textit{Includes}, \textit{Is\_Included} and \textit{Simultaneous} are all less than 4\%.  

\section{Related Work}

Early studies on temporal relation extraction mainly model it as a pairwise classification problem~\cite{mani2006machine,verhagen2007semeval,verhagen2008temporal,verhagen2010semeval,bethard2016semeval,macavaney2017guir} and rely on hand-crafted features and rules~\cite{verhagen2008temporal,4338327} 
to extract temporal event relations. Recently, deep neural networks ~\cite{dligach2017neural,tourille2017neural} and large-scale pre-trained language models~\cite{han2019deep,han2020deer,wang2020joint,zhou2020clinical} are further employed and show state-of-the-art performance.

Similar to our approach, several studies~\cite{ling2010temporal,nikfarjam2013towards,mirza2016catena,meng-etal-2017-temporal,cheng2017classifying,huang2017improving} also explored syntactic path between two events for temporal relation extraction. 
Different from previous work, our approach considers three important sources of temporal cues: \textit{local context}, denoting the neighbors of each event node within the dependency graph; \textit{connection of two event mentions}, which is based on the dependency path between two event mentions; and \textit{rich semantics of concepts and dependency relations}, for example, the dependency relation \textit{nmod} between two event mentions usually indicates a \tmprel{Before} relationship. All these indicative features are automatically selected and aggregated with the multi-head self-attention and our new syntax-guided attention mechanism.

Our work is also related to the variants of Graph Neural Networks (GNN)~\cite{kipf2016semi,velivckovic2017graph,zhou2018graph}, especially Graph Transformer~\cite{yun2019graph,chen2019path,hu2020heterogeneous,wang2020amr}. Different from previous GNNs which aim to capture the context from neighbors of each node within the graph, in our task, we aim to select and capture the most meaningful temporal cues for two event mentions from their connections within the graph as well as their surrounding contexts.


\section{Conclusion}
Temporal relationship between events is important for understanding stories described in natural language text, and a main challenge is how to discover and make use of the connection between two event mentions, especially when the event pair is far apart in text. This paper proposes a novel Syntax-guided Graph Transformer (\this{}) that represents the connection between an event pair via additional attention heads over dependency parsing trees. Experiments on benchmarking datasets, \matres{}, \tbd{}, and a contrast set of \matres{}, show that our approach significantly outperforms previous state-of-the-art methods in a variety of settings, including event detection, temporal relation classification (where events are given), and temporal relation extraction (where events are predicted). In the future, we will investigate the potential of this approach to other relation extraction tasks.

\section*{Acknowledgements}
We thank the anonymous reviewers and area chair for their valuable time and constructive comments. We also thank the support from the Amazon Research Awards.

\bibliographystyle{acl_natbib}
\bibliography{naacl22}

\end{document}